\documentclass[11pt]{article}
\usepackage{coling2020}
\usepackage{times}
\usepackage{url}
\usepackage{latexsym}

\usepackage{graphicx}
\usepackage{CJK}
\usepackage{multirow}
\usepackage{amsmath}
\usepackage{amsthm}
\usepackage{amssymb}
\usepackage{booktabs}
\usepackage{algorithm}
\usepackage{algorithmic}
\usepackage{paralist} 
\usepackage{enumitem}
\usepackage{xcolor}

\definecolor{RANK1}{HTML}{DA2020}
\definecolor{RANK2}{HTML}{E74C3C}
\definecolor{RANK3}{HTML}{F1948A}
\definecolor{RANK4}{HTML}{F5B7B1}

\colingfinalcopy 

\title{Synonym Knowledge Enhanced Reader for Chinese Idiom Reading Comprehension}

\author{Siyu Long$^{1*}$, Ran Wang$^{1*}$, Kun Tao$^{1}$, Jiali Zeng$^{2}$, Xin-Yu Dai$^{1}$ \\
  $^{1}$National Key Laboratory for Novel Software Technology, Nanjing University \\
  $^{2}$Smart Platform Product Department of Tencent Inc., China\\
  {\tt {\{longsy, wangr, taok\}}@smail.nju.edu.cn}\\
  {\tt lemonzeng@tencent.com, daixinyu@nju.edu.cn}

\\}

\date{}

\begin{document}
\maketitle
\begin{abstract}

Machine reading comprehension (MRC) is the task that asks a machine to answer questions based on a given context. For Chinese MRC, due to the non-literal and non-compositional semantic characteristics, Chinese idioms pose unique challenges for machines to understand. Previous studies tend to treat idioms separately without fully exploiting the relationship among them. In this paper, we first define the concept of literal meaning coverage to measure the consistency between semantics and literal meanings for Chinese idioms. With the definition, we prove that the literal meanings of many idioms are far from their semantics, and we also verify that the synonymic relationship can mitigate this inconsistency, which would be beneficial for idiom comprehension. Furthermore, to fully utilize the synonymic relationship, we propose the synonym knowledge enhanced reader. Specifically, for each idiom, we first construct a synonym graph according to the annotations from a high-quality synonym dictionary or the cosine similarity between the pre-trained idiom embeddings and then incorporate the graph attention network and gate mechanism to encode the graph. Experimental results on ChID, a large-scale Chinese idiom reading comprehension dataset, show that our model achieves state-of-the-art performance.\footnote{Our code is available at \url{https://github.com/njunlp/SKER}} 


\end{abstract}

\section{Introduction}
\label{intro}

%
%
\blfootnote{
    %
    %
    %
    %
    %
    %
    *\ Equal contribution.\\
    This work is licensed under a Creative Commons 
    Attribution 4.0 International License.
    License details:
    \url{http://creativecommons.org/licenses/by/4.0/}.
}

Machine reading comprehension is the task that asks a machine to answer questions based on a given context. Although various advanced neural models have been proposed in recent years \cite{HermannKGEKSB15,ChenBM16}. Chinese reading comprehension is still a challenging task \cite{HeLLLZXLWWSLWW18,SunYYC20}. One of the reasons is the existence of Chinese idioms, also known as ``chengyu", whose non-literal and non-compositional characteristics make them difficult for machines to understand \cite{blacklist,chid}.

Recently, some studies \cite{JiangZHJ18,chid} claimed that good idiom representations are essential for idiom comprehension. Meanwhile, many methods have been proposed to obtain idiom representations, which can be divided into two categories: First, a straightforward way is to treat each Chinese idiom as a single token and obtain its representation by training on a large-scale corpus \cite{SongSLZ18,chid}. However, compared to common words, Chinese idioms appear less frequently. These methods are likely not sufficient to learn good enough idiom representations \cite{BahdanauBJGVB17}; Second, since most Chinese idioms consist of four Chinese characters, another direct approach is to treat each Chinese idiom as a regular multi-word expression. In this category, most studies combine the distributional word embeddings \cite{word2vec,word2vec2} to learn representations of multi-word expressions. Additionally, some methods try to enhance the representations by employing a more powerful compositionality function \cite{GrefenstetteS11,Blacoe,SocherPWCMNP13}, integrating combination rules \cite{KoberWRW16,WeirWRK16}, or incorporating the general linguistic knowledge \cite{QiHYLCLS19}. By combining the word embeddings, we can alleviate the sparsity of idioms. Nevertheless, the semantics of idioms are often different from their literal meanings \cite{blacklist,chid}, making them difficult to understand only by the composition of Chinese characters.

\begin{figure}[!tb]
    \centering
    \begin{minipage}[t]{0.45\textwidth}
        \centering
        \includegraphics[width=0.75\textwidth]{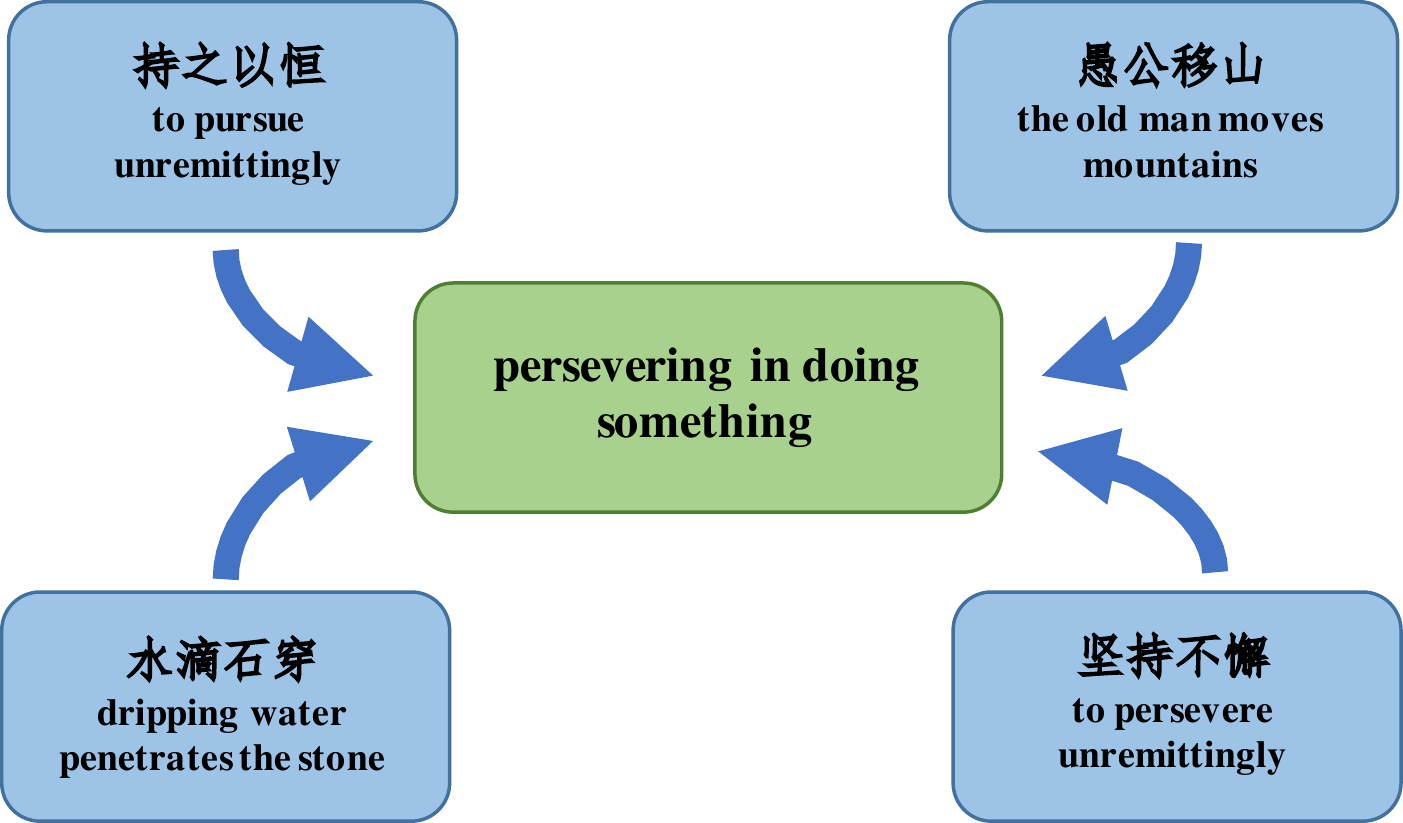}
        \caption{Example of synonymic relationship among Chinese idioms. Here, four Chinese idioms share similar semantics, but express in different ways.}
        \label{fig:ex}
    \end{minipage}
    \quad
    \begin{minipage}[t]{0.45\textwidth}
        \centering
        \includegraphics[width=0.55\textwidth]{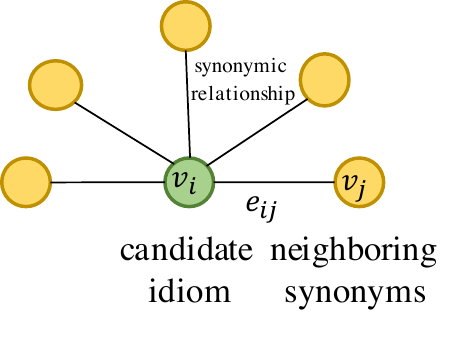}
        \caption{Example of synonym graph. The green node is the candidate idiom and yellow nodes are the synonyms. The edges in the graph represent the synonymic relationship.}
	    \label{fig:placeholder}
    \end{minipage}
\end{figure}

Notably, the above works ignore the relationship among the Chinese idioms. Intuitively, there exists a general relationship that can be used in representing Chinese idioms, e.g., the synonymic relationship \cite{chid}. In this paper, we argue that this relationship is beneficial to getting better representations and for Chinese idiom comprehension. Taking Figure \ref{fig:ex} for example, four Chinese idioms are sharing similar meanings: \textit{persevering in doing something}. Among them, the semantics of ``\begin{CJK*}{UTF8}{gbsn}坚持不懈\end{CJK*}" is the same as its literal meaning, i.e., \textit{to persevere unremittingly}. But ``\begin{CJK*}{UTF8}{gbsn}水滴石穿\end{CJK*}" (literal meaning: \textit{dripping water penetrates the stone}) expresses it by using a metaphor. Instinctively, we may be able to acquire the semantics of ``\begin{CJK*}{UTF8}{gbsn}水滴石穿\end{CJK*}" more efficiently through its synonymic relationship with ``\begin{CJK*}{UTF8}{gbsn}坚持不懈\end{CJK*}". To verify this, in this paper, we first define a concept of literal meaning coverage to measure the consistency between semantics and literal meanings for Chinese idioms. Then we conduct quantitative experiments to analyze the coverage on our sampled idioms. The experimental results show that the literal meanings of many idioms are far from their semantics, and the synonymic relationship can mitigate this inconsistency. Specifically, the semantics of 41.5\% sampled idioms are more accessible with the help of synonymic relationship, which suggests that the use of synonymic relationship is a potential auxiliary for idiom comprehension.

Based on the discovery above, we propose the synonym knowledge enhanced reader (SKER) to fully utilize the synonymic relationship. Concretely, for each idiom, according to the annotations from a high-quality synonym dictionary or the cosine similarity between the pre-trained idiom embeddings, we construct a synonym graph to represent the synonymic relationship and then incorporate the graph attention network and gate mechanism to encode the graph. We evaluate our model on ChID, a large-scale Chinese idiom reading comprehension dataset, and experimental results show that it achieves state-of-the-art performance. 

Our contributions are threefold: (1) We quantitatively prove the inconsistency between semantics and literal meanings of Chinese idioms; (2) We verify the synonymic relationship among idioms can mitigate this inconsistency effectively; (3) We propose a novel model to fully utilize the relationship and achieve state-of-the-art performance in ChID, a large-scale Chinese idiom reading comprehension dataset.


\section{Preliminary Study of Synonymic Relationship}
In this section, we first define the concept of literal meaning coverage for Chinese idioms. Then quantitative analyses are conducted to present the inconsistency between literal meanings and semantics of Chinese idioms and the effectiveness of synonymic relationship in mitigating this inconsistency.


\subsection{Literal Meaning Coverage for Chinese Idioms}

Although Chinese idioms are characterized in terms of non-literal and non-compositional \cite{blacklist,chid}, not every idiom's semantics is inaccessible through its literal meaning. Different idioms show varying degrees of consistency between their literal meanings and semantics. To measure the consistency, we define the coverage of literal meaning to semantics as literal meaning coverage (LMC). Specifically, according to the different situations between literal meanings and semantics, we define three LMCs denoted by 1, 2, and 3, where larger numbers mean better consistency. An example of each LMC is shown in Table \ref{tab:scd}. Next, we give a brief explanation for these LMCs:


\begin{itemize}[itemsep= 0.5 pt,topsep = 5 pt]
    \item LMC 1: When the literal meaning of a Chinese idiom is entirely different from its semantics, especially when the semantics is highly dependent on metaphor, the LMC of this idiom equals 1.
    \item LMC 2: When the semantics of a Chinese idiom is partially covered by its literal meaning, the LMC of this idiom equals 2.
    \item LMC 3: When the literal meaning of a Chinese idiom is identical to its semantics, the LMC of this idiom equals 3.
\end{itemize}

\begin{table}[hbt]
\centering
\begin{tabular}{ccll}
\toprule
LMC &
  Chinese Idiom &
  \multicolumn{1}{c}{Literal Meaning} &
  \multicolumn{1}{c}{Semantics} \\
\midrule
1 &
  \begin{CJK*}{UTF8}{gbsn}水滴石穿\end{CJK*} &
  \begin{tabular}[c]{@{}l@{}}dripping water penetrates the stone.\end{tabular} &
  constant perseverance yields success. \\
\midrule
2 &
  \begin{CJK*}{UTF8}{gbsn}一泻千里\end{CJK*} &
  (of a river) rushing along. &
  \begin{tabular}[c]{@{}l@{}} (of a river) rushing along.\\ 
  a bold and flowing calligraphic style.
  \end{tabular} \\
\midrule
3 &
  \begin{CJK*}{UTF8}{gbsn}坚持不懈\end{CJK*} &
  \begin{tabular}[c]{@{}l@{}}to persevere unremittingly.\end{tabular} &
  \begin{tabular}[c]{@{}l@{}}persevering in doing something.\end{tabular} \\
  \bottomrule
\end{tabular}
\caption{
Different literal meaning coverage and corresponding examples.
}
\label{tab:scd}
\end{table}

Based on the above definition, we can give a quantitative analysis concerning the consistency between literal meanings and semantics of Chinese idioms. We randomly sample 200 idioms from Baidu Hanyu\footnote{\url{https://dict.baidu.com/}}, an online idioms dictionary, and for each idiom, three native speakers are asked to label the LMC. We take the average of the 3 LMCs as the final LMC, a real number between 1 and 3. 
During the labeling, the dictionary definition of each idiom is available for the native speakers.  For comparison, we also randomly sample 200 common words from HSK vocabulary\footnote{\url{http://www.chinesetest.cn/godownload.do}}, which aims for primary Chinese teaching and ask the native speakers to label their LMCs. As shown in Table \ref{tab:lmc}, there are 66\% idioms whose final LMCs are lower than 2.33 while the corresponding percentage of common words is 1.5\%, which indicates that the literal meanings of many Chinese idioms are far from their semantics and also suggests that the methods which combine the characters are inappropriate for Chinese idioms. The Fleiss' kappas \cite{fleiss1971measuring} of idioms and common words are 0.655 and 0.833, indicating substantial and almost perfect agreement among annotators.

\subsection{Mitigating the Inconsistency with Synonymic Relationship}
\label{sec:miti}

The results mentioned above show that the literal meanings of many idioms are inconsistent with their semantics, but we verify that this inconsistency can be mitigated with the synonymic relationship. For previously sampled 200 idioms, we find their synonyms by means of a high-quality synonym dictionary\footnote{\url{https://github.com/Keson96/SynoCN}}, which contains the annotations of synonymic relationship and is also obtained from Baidu Hanyu. Similar to the idioms, each synonym is also labeled by three native speakers and the average of LMCs is also the final LMC of the synonym. We compare the final LMCs of idioms and the average of the final LMCs of their synonyms to check the change of final LMCs. The results are reported in Table \ref{tab:change}, which show that 41.5\% sampled idioms can have higher final LMCs with the assistance of synonymic relationship, exhibiting its effectiveness in mitigating the inconsistency. Accordingly, the mitigation can potentially ease the machine access to the semantics and is conductive to idiom comprehension. However, there are also 25.5\% idioms that have lower final LMCs, which suggests the necessity of a noise reduction process when leveraging the synonymic information.



\begin{minipage}{\textwidth}
\centering
 \begin{minipage}[t]{0.45\textwidth}
    \begin{tabular}{cccc}
		\toprule
		Final LMC    &  Idioms &  Common Words  \\
		\midrule
		{[}1.00, 1.66) & 45.5\%  & 0.0\%  \\
		{[}1.66, 2.33) & 20.5\%  & 1.5\%  \\
		{[}2.33, 3.00] & 34.0\%  & 98.5\% \\
		\bottomrule
	\end{tabular}
	\makeatletter\def\@captype{table}\makeatother\caption{
	    Distribution of final literal meaning coverage on different Chinese elements. 
	}
	\label{tab:lmc}
  \end{minipage}
\quad
  \begin{minipage}[t]{0.45\textwidth}
   \centering
	\begin{tabular}{cccc}
		\toprule
		Final LMC Change    & Proportion &     \\
		\midrule
		Higher & 41.5\%    \\
		Equal & 33.0\%  \\
		Lower & 25.5\%  \\
		\bottomrule
	\end{tabular}
	  \makeatletter\def\@captype{table}\makeatother\caption{Change of final literal meaning coverage.}
	\label{tab:change}
   \end{minipage}
\end{minipage}

\section{Proposed Method}


In this section, we first give a formulation of Chinese idiom reading comprehension. Then we introduce how to construct the synonym graph. Finally, we propose the synonym knowledge enhanced reader (SKER) with the synonym graph encoding.

\subsection{Formulation}


Each instance in Chinese idiom reading comprehension contains a passage $P = [w_1, \cdots,w_b,\cdots, w_n]$ with a blank $w_b$, which is used to replace the idiom and a set of candidate idioms $Z = \{z_1, z_2, \cdots, z_m\}$. Our goal is to choose the correct idiom from the candidate set $Z$ to fill up the blank $w_b$.


\subsection{Construction of Synonym Graph}

For each idiom in the candidate set, in order to represent the synonymic relationship, we construct a synonym graph $\mathcal{G} = (\mathcal{V}, \mathcal{E})$, where the node set $\mathcal{V}$ is the set of the candidate idiom and its neighboring synonyms, and the edge set $\mathcal{E}$ is the synonymic relationship among them. We construct the synonym graph according to the aforementioned synonym dictionary, which contains high-quality annotations of synonymic relationship. Specifically, take Figure \ref{fig:placeholder} for example, for node $v_i$ and $v_j$, an undirected edge $e_{ij}$ is added to the graph when we can find their corresponding idioms are annotated as synonyms in the dictionary. As we mainly focus on the enhancement of candidate idioms, the relationship among neighboring synonyms is ignored.

However, building a high-quality dictionary is labor-intensive, and \newcite{word2vec} claimed that the semantic relevance between two words can be measured by the cosine similarity of their embeddings. Therefore, we also construct the graph with pre-trained embeddings. Similarly, for node $v_i$ and $v_j$, an undirected edge $e_{ij}$ is added to the graph when the cosine similarity of their corresponding idiom embeddings is higher than a pre-defined threshold. As \newcite{chid} claimed that there is a substantial probability that two idioms are near-synonyms when their cosine similarity is higher than 0.65 in ChID, which is the dataset we used in this work, the value of the threshold is set as 0.65.

\subsection{Synonym Knowledge Enhanced Reader}
\begin{figure}[hbt]
	\centering
	\includegraphics[width=0.75\textwidth]{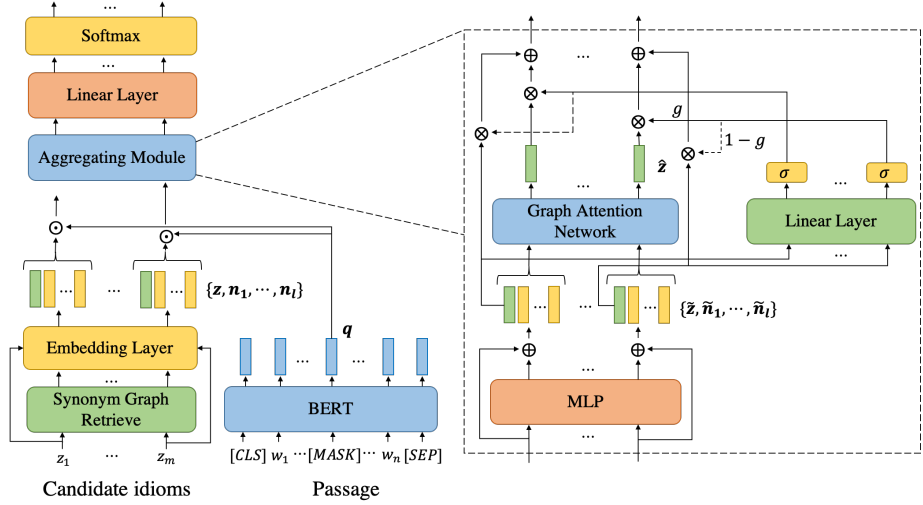}
	\caption{
		Architecture of synonym knowledge enhanced reader. $\odot$ is element-wise product. $\oplus$ is element-wise addition. $\otimes$ is scalar multiplication of vectors.
	}
	\label{fig:overview}
\end{figure}

\paragraph{Overview}


Figure \ref{fig:overview} gives an overview of our proposed model. 
The key challenge for better idiom representations and answer predictions is to utilize the synonym graph while reducing the potential noise effectively. Facing the challenge, we propose our model, which is composed of an encoding module, an aggregating module, and a predicting module. Specifically, in the aggregating module, we adopt the graph attention network to fully utilize the synonym graph and meanwhile filter the potential noise with the gate mechanism.


\paragraph{Encoding Module}

We utilize BERT \cite{DevlinCLT19}, which has shown marvelous performance in various NLP tasks, to encode passages. Concretely, we first tokenize a passage with the WordPiece vocabulary, which is used in the original paper of BERT, and then generate the input sequence by concatenating the \texttt{[CLS]} token, the tokenized passage, and the \texttt{[SEP]} token, where \texttt{[CLS]} and \texttt{[SEP]} are two special tokens used in BERT for leading and ending the sequence. The blank in the passage is replaced by the \texttt{[MASK]} token. Through BERT, the passage is encoded into a list of word representations. Here, we omit a detailed description of BERT and refer interested readers to \newcite{DevlinCLT19}. We take the hidden state of the blank as the final representation of passage $\mathbf{q}\in\mathbb{R}^{d}$, where $d$ is the hidden state size.


Given the corresponding candidate Chinese idioms set $Z = \{z_1, z_2, \cdots, z_{m}\}$ and their synonym graphs $\mathcal{G}$s, we first retrieve the synonyms of each candidate. For candidate $z_i$, let $\mathcal{N}_{z_{i}} = \{
n_1^{(z_i)}, n_2^{(z_i)}, \cdots, n_{l_{z_i}}^{(z_i)}
\}$ denote the synonyms, where $l_{z_i}$ is the cardinality of $\mathcal{N}_{z_{i}}$. Following \newcite{chid}, we use a embedding layer to convert the candidate and its synonyms into a set of low-dimensional vectors $\{
\mathbf{z}_i,
\mathbf{n}_{1}^{(z_{i})},
\cdots,
\mathbf{n}_{l_{z_{i}}}^{(z_{i})}
\}$. For brevity, we eliminate part of superscripts and subscripts, and use $\mathbf{z}$ and $\{\mathbf{n}_1, \cdots, \mathbf{n}_{l}\}$ to denote the vectors of the candidate idiom and its synonyms respectively. 


To allow the candidate idiom to perceive the information of the passage, we update encoded idiom representation $\mathbf{z} \in \mathbb{R}^{d}$ as follow:

\begin{equation}
\mathbf{z}_{t} = \mathbf{q} \odot \mathbf{z}
\end{equation}
where $\odot$ denotes the element-wise product. Here, we can also employ other interaction methods, which is reserved for further exploration. In the same way, the corresponding synonym $\mathbf{n}$ also interact with the passage representation, which is denoted by $\mathbf{n}_{t}$.

\paragraph{Aggregating Module}
As mentioned in section \ref{sec:miti}, the synonymic relationship among idioms is probably advantageous for idiom comprehension. To fully utilize the relationship, we employ a graph attention network to aggregate the representations of candidate idioms and synonyms. First, we use a multilayer perceptron to update the passage-aware representations $\mathbf{z}_t$ and $
\{
\mathbf{n}_{t1}, 
\mathbf{n}_{t2},
\cdots,
\mathbf{n}_{tl}
\}
$.
\begin{equation}
\label{equ:transform}
\tilde{\mathbf{z}} = 
\mathbf{W}^{\top}_{s}(\mathrm{ReLU}(\mathbf{W}^{\top}_{t}\mathbf{z}_{t}+\mathbf{b}_{t})) + \mathbf{b}_{s} + \mathbf{z}_{t}
\end{equation}
where $\mathbf{W}_{t}, \mathbf{W}_{s} \in\mathbb{R}^{d\times d}$ are trainable weight matrices and $\mathbf{b}_t, \mathbf{b}_s \in \mathbb{R}^{d}$ are trainable biases. 
Similarly, each updated representation of synonym $\tilde{\mathbf{n}}$ is also obtained from the same perceptron, and we concatenate these transformed representations as follows:
\begin{equation}
\mathbf{N} = \parallel \{\tilde{\mathbf{z}}, \tilde{\mathbf{n}}_{1}, \cdots, \tilde{\mathbf{n}}_{l}\} \in \mathbb{R}^{d\times (l+1)}
\end{equation}
where $\parallel$ denotes the vector concatenation operation.


Different Chinese idioms in the same synonym group may not be equally importance for answering different questions and as mentioned in section \ref{sec:miti}, the synonym graph is likely to contain noise. Graph attention network (GAT) presented by \newcite{VelickovicCCRLB18} can assign different weights to different nodes, which can be used to model the different significance of neighboring synonyms and to reduce the noise on the graph. Therefore, we employ a graph attention network to fuse the information of neighboring synonyms into their corresponding candidate idiom. 

At first, we split the matrix $\mathbf{N}$ into $h\in\mathbb{N}$ sub-matrices $\mathbf{N}^{(1)}, \cdots, \mathbf{N}^{(h)} \in \mathbb{R}^{ {d_h}\times {(l+1)}}$.
Here, $d_h=\frac{d}{h} \in \mathbb{N}$ is the dimensions of sub-matrices and $h$ is the number of attention heads.
Similarly, $\tilde{\mathbf{z}}$ is converted into vectors $\tilde{\mathbf{z}}^{(1)}, \cdots, \tilde{\mathbf{z}}^{(h)} \in \mathbb{R}^{ {d_h}}$.
Considering varying significance of different Chinese idioms, different weights are assigned to them in this model. This process can be define as:

\begin{equation}
\mathrm{head}_i
=
\mathrm{softmax}(
\frac{
        (\mathbf{W}_{i}^{Q}\tilde{\mathbf{z}}^{(i)})^{\top}
        \mathbf{W}_{i}^{K} \mathbf{N}^{(i)}
        }{\sqrt{d_{h}}}
)
\big(\mathbf{W}_{i}^{V} \mathbf{N}^{(i)}\big)^{\top}
\end{equation}
where $\mathbf{W}_{i}^{Q}, \mathbf{W}^{K}_{i}, \mathbf{W}_{i}^{V} \in \mathbb{R}^{d_h\times d_h}$ are trainable weight matrix.

Then we collect the information of different attention heads and update the candidate idiom representation as follows:
\begin{equation}
\hat{\mathbf{z}} = \mathbf{W}^{O} (\parallel_{i=1}^{h} \mathrm{head}_i)^{\top} + \mathbf{b}^{O}
\end{equation}
where $\mathbf{W}^{O}\in\mathbb{R}^{d\times d}$ is a learnable weight matrix and $\mathbf{b}^{O}\in\mathbb{R}^{d}$ is the bias.


In order to further reduce the potential impact of synonym graph noise, we use the gate mechanism to control the fusion process of the synonym knowledge enhanced representation $\hat{\mathbf{z}}$ and the passage-aware representation $\tilde{\mathbf{z}}$.
Specifically, we get the final representation of candidate idiom $\check{\mathbf {z}} $ as follows:

\begin{equation}
    \begin{split}
        g &= \sigma(\mathbf{W}_{g} \tilde{\mathbf{z}} + \mathbf{b}_{g})\\
        \check{\mathbf{z}} &= g\cdot\hat{\mathbf{z}} + (1-g)\cdot\tilde{\mathbf{z}}\\
    \end{split}
\end{equation}
where $\mathbf{W}_g \in \mathbf{R}^{d\times d}$ and $\mathbf{b}_{g} \in \mathbb{R}^{d}$ are trained parameters of a linear layer.

\paragraph{Predicting Module}
Finally, the representation of all candidates $\check{\mathbf{Z}} = \{\check{\mathbf{z}}_1, \check{\mathbf{z}}_2, \cdots, \check{\mathbf{z}}_{m}\}$ are used to predict the answer.

\begin{equation}
    \Pr(y_i|P) = \mathrm{softmax}_{i}(\mathbf{u}^{\top}_{o}\check{\mathbf{Z}} + \mathbf{b}_o)
\end{equation}
where $\mathbf{u}_{o} \in \mathbb{R}^{d}$, $\mathbf{b}_o \in \mathbf{R}$ are parameters mapping the candidate representation into a score.

The goal of training is to minimize the cross-entropy loss between ground truths and predictions:
\begin{equation}
    \mathrm{loss} = -y_i\mathrm{log}\Pr(y_i|P)
\end{equation}
where $y_i$ is the gold answer.

\section{Experiments}
\subsection{Dataset}
We evaluate our model on ChID \cite{chid}, a large-scale idiom reading comprehension dataset. In this dataset, idioms in passages are replaced by blank symbols, and in order to complete the passages, the correct idioms need to be chosen from well-designed candidates. The passages in ChID are collected from novels and essays on the Internet and news articles provided by \newcite{thuctc}. To assess the generalization ability of models, news and novels are treated as in-domain data, while essays are reserved for \textbf{Out} test. We also evaluate our models on two additional test sets: \textbf{Ran} and \textbf{Sim}. Both datasets have the same passages with \textbf{Test}, but candidate idioms are designed differently. In \textbf{Ran}, all the candidates are sampled randomly; in \textbf{Sim}, the candidate idioms are sampled from the top 10 similar idioms. The detailed statistics of this dataset are summarized in Table \ref{tab:data-stats}.


\subsection{Compared Methods}
We evaluate and compare our method with four competitive baseline models: \textbf{Language Model} (LM) utilizes a bidirectional LSTM to obtain the representations of passages, and use them to score candidate Chinese idioms; \textbf{Attentive Reader} (AR) augments the bidirectional LSTM with the attention mechanism to attentively read the context; \textbf{Stanford Attentive Reader} (SAR) applies a bilinear matrix when producing context-aware passage representations; \textbf{BERT} instead employs powerful bidirectional encoder's representations from transformers as the hidden state of passages.

As \newcite{JiangZHJ18}, \newcite{chid} claimed that idiom representations are essential for idiom comprehension. We also evaluate the performance of four different ways to represent idioms: in \textbf{ran}, we simply represent Chinese idioms with randomly initialized vectors; in \textbf{pre}, we replace the randomly initialized idiom vectors with word vectors trained on a large-corpus; in \textbf{text}, we obtain the idiom representations through their Chinese characters' embeddings. Specifically, we treat each Chinese idiom as a sequence of four Chinese characters. Then we average the pre-trained embeddings of these characters, and finally, the idiom representation is obtained through a layer of feedforward neural network; in \textbf{def}, we crawl the definitions of each idiom from the Internet, encode the definitions through BERT, and then feed the average of the encoded results into a feedforward neural network, and the outputs are used as the final representations of Chinese idioms.   

\subsection{Implemented Details}

For LM, AR and SAR, we mostly follow the same architecture, optimization, and data pre-processing used in \newcite{chid}. For models based  on BERT, we follow \newcite{BERT-wwm} to use LTP for Chinese word segmentation and implement them with AllenNLP \cite{allennlp}. Instead of using the official BERT (Chinese), we employ an open source BERT with whole word masking\footnote{
	\url{https://github.com/ymcui/Chinese-BERT-wwm}
} as the pre-trained encoder, which is more effective in handing Chinese texts. We use the embeddings from \newcite{SongSLZ18} to construct the Chinese idiom synonym graph. The random idiom embeddings are initialized according to \newcite{xavier_uniform} and the size of embeddings is 200. For BERT-based models, there is an extra matrix that can convert the above idiom embeddings into 768-dimensional vectors, which is the hidden size of BERT. Adam \cite{adam} is used to optimize all the models with the initial learning rate 2e-5. The numbers of attention head, dropout rate and batch size are 2, 0.2 and 128. All hyper-parameters are tuned on the \textbf{Dev} dataset, and the training is stopped when the accuracy on \textbf{Dev} does not improve within one epoch.

\section{Results and Discussion}
\subsection{Experimental Results}
The results of our model and compared models are reported in Table \ref{tab:main-results}. The experimental results reveal that:


\begin{minipage}{\textwidth}
\begin{center}
\begin{minipage}[t]{0.45\textwidth}
\vspace{40pt}
\scriptsize
    \begin{tabular}{lrrrrr}
	\toprule
	& Train   & Dev    & Test   & Out    & Total   \\
	\midrule
	\# document         & ~521k & 20k & 20k & 20k & ~581k \\
	average \# token per doc & 99      & 99     & 99     & 127    & 100     \\
	average \# blank per doc& 1.25    & 1.24   & 1.25   & 1.49   & 1.25 \\
	\midrule
	\# idiom & 3.8k   & 3.5k  & 3.5k  & 3.6k  & 3.8k   \\
	average idiom frequency   & 168.6   & 7.2    & 7.1    & 8.3    & 189.6   \\
	\bottomrule
\end{tabular}
\makeatletter\def\@captype{table}\makeatother\caption{
	ChID dataset statistics.}

\label{tab:data-stats}

\hspace*{\fill}

\hspace*{\fill}

\hspace*{\fill}

\centering
\begin{tabular}{llllllll}
\toprule
\multicolumn{1}{c}{Model} & Dev  & Test & Ran  & Sim  & Out  & Ave & delta \\
\midrule
SKER                      & 76.0 & 76.1 & 87.0 & 68.6 & 68.3 & 75.0 & - \\
w/o synonym               & 75.2 & 75.5 & 84.6 & 67.4 & 67.2 & 73.7 & \textbf{-1.3}\\
w/o gate                    & 75.5 & 75.6 & 86.6 & 68.1 & 67.5 & 74.4 & \textbf{-0.6} \\
w/o gate \& GAT              & 74.7 & 74.6 & 85.5 & 67.0 & 66.2 & 73.3 & \textbf{-1.7} \\
\bottomrule
\end{tabular}
	\makeatletter\def\@captype{table}\makeatother\caption{
		Ablation study result.
	}
	\label{tab:ablation-results}
\end{minipage}
\qquad \quad
\begin{minipage}[t]{0.45\textwidth}
\vspace{0pt}
\scriptsize
\begin{tabular}{clllllll}
\toprule
\multicolumn{2}{c}{Model}      & Dev  & Test & Ran  & Sim  & Out  & Ave  \\
\midrule
\multirow{4}{*}{LM}   & ran  & 69.3 & 69.1 & 77.1 & 63.6 & 59.6 & 67.4 \\
                      & pre  & 71.6 & 71.2 & 80.4 & 65.0 & 61.5 & 69.5 \\
                      & text & 68.1 & 67.7 & 76.6 & 61.6 & 57.0 & 65.7 \\
                      & def  & 69.8 & 69.3 & 78.5 & 62.9 & 60.7 & 67.9 \\
\midrule
\multirow{4}{*}{AR}   & ran  & 69.6 & 69.7 & 77.9 & 63.6 & 58.8 & 67.5 \\
                      & pre  & 72.6 & 72.5 & 81.7 & 66.0 & 63.3 & 70.9 \\
                      & text & 67.3 & 66.8 & 76.2 & 60.8 & 57.0 & 65.2 \\
                      & def  & 68.7 & 68.9 & 77.9 & 62.6 & 59.0 & 67.1 \\
\midrule
\multirow{4}{*}{SAR}  & ran  & 69.3 & 69.1 & 77.5 & 63.2 & 59.1 & 67.2 \\
                      & pre  & 71.2 & 71.1 & 80.7 & 65.0 & 61.6 & 69.6 \\
                      & text & 66.7 & 66.2 & 75.5 & 60.3 & 55.8 & 64.5 \\
                      & def  & 69.8 & 68.7 & 78.1 & 62.4 & 58.5 & 66.9 \\
\midrule
\multirow{4}{*}{BERT} & ran  & 65.8 & 66.1 & 73.9 & 61.1 & 55.9 & 64.3 \\
                      & pre  & 74.7 & 74.6 & 85.5 & 67.0 & 66.2 & 73.3 \\
                      & text & 71.0 & 71.5 & 81.0 & 65.2 & 63.0 & 70.2 \\
                      & def  & 62.9 & 62.9 & 76.1 & 55.3 & 55.6 & 62.5 \\
\midrule
\multirow{2}{*}{SKER} & ran  & 74.7 & 74.9 & 84.8 & 68.2 & 66.2 & 73.5 \\
 & pre & \textbf{76.0} & 76.1 & \textbf{87.0} & 68.6 & \textbf{68.3} & 75.0 \\
\midrule
\multirow{2}{*}{SKER*} & ran  & 74.6 & 74.6 & 84.3 & 68.1 & 66.0 & 73.3 \\
 & pre & 75.9 & \textbf{76.3} & 86.9 & \textbf{68.8} & 68.2 & \textbf{75.1} \\
 \bottomrule
\end{tabular}
\makeatletter\def\@captype{table}\makeatother\caption{
 Model comparison result (accuracy, \%). ``Ave" denotes the averaged accuracy over Test, Ran, Sim and Out. 
 The best results are in bold.
}
\label{tab:main-results}
\end{minipage}
\end{center}
\end{minipage}

First, synonymic relationship among idioms can significantly improve the performance of models. After employing the synonymic relationship, the performance of SKER-ran can get an absolute improvement of 8.7\% on average comparing with BERT-ran; even compared with BERT-pre, SKER-pre can still obtain a 1.7\% absolution improvement.



Second, the difficulty of obtaining Chinese idioms' semantics through their literal texts is definitively proved again by comparing the performances of all the non-BERT-based ``text" baselines with their corresponding ``ran" and  ``pre" models. 



Third, high-quality Chinese idiom embeddings can improve the performance. All the models exploiting the pre-trained embeddings have higher accuracy than the models which only represent the idioms with random vectors. This result is consistent with the findings of previous works \cite{chid}. It shows that the quality of idiom embeddings largely determines the performance of the model on the reading comprehension task.



Finally, the source of the synonymic relationship has little effect on the performance of SKER. Specifically, the performance of SKER*, whose relationship is from high-quality synonym dictionary, is only 0.1\% higher than SKER, whose relationship is from pre-trained embeddings. 

\subsection{Ablation Study}


To further evaluate the contribution of each part in our proposed model, three variants of SKER are provided. The notations of ``w/o synonym", ``w/o gate" and ``w/o gate \& GAT" denote the models, which replace the synonyms with random idioms, remove gate mechanism and discard both graph attention network and gate mechanism respectively.

The results of the ablation study in Table \ref{tab:ablation-results} reveal that these three parts are helpful for idiom reading comprehension. 
It verifies the significance of the synonymic relationship among idioms.
The gate mechanism is also beneficial to performance improvement because it can reduce the noise mentioned in section \ref{sec:miti}.
Notably, with random idioms, our model can still obtain a slight improvement compared to BERT. The reason is that the random idioms can still contain synonyms which our model can capture keenly.




\subsection{Case Study}
Here, we give an example of idiom prediction with the synonymic relationship, to show that our model can correctly utilize the relationship.

\begin{table}[htb]
\centering
\begin{tabular}{l|l}
\toprule
Passage &
\begin{tabular}[c]{@{}l@{}}
     \begin{CJK*}{UTF8}{gbsn}最惨痛的当然是主教练姜正秀，不光电脑被偷，衣服、鞋子\end{CJK*}\\
     \begin{CJK*}{UTF8}{gbsn}也被毛贼 \_\_\_，以至于他不得不穿着拖鞋赶路。\end{CJK*}\\
     
    The most painful thing is that the head coach Jiang Zhengxiu, not only \\
    the computer was stolen, but his clothes and shoes were also \_\_\_ ,
    so that\\ he had to wear slippers to get off the road.
\end{tabular} \\ \midrule
Ground Truth &

\begin{tabular}[c]{@{}l@{}}
\begin{CJK*}{UTF8}{gbsn}顺手牵羊\end{CJK*}\\
literal meaning: walk off with a sheep\\
semantics: steal something in passing
\end{tabular} \\ \midrule

Attention Head - 1 & {\color{RANK1} \begin{CJK*}{UTF8}{gbsn}见财起意\end{CJK*} (0.387)}, {\color{RANK3} \begin{CJK*}{UTF8}{gbsn}物归原主\end{CJK*} (0.172)}, {\color{RANK2} \begin{CJK*}{UTF8}{gbsn}据为己有\end{CJK*} (0.319)}, {\color{RANK4} \begin{CJK*}{UTF8}{gbsn}偷梁换柱\end{CJK*} (0.122)} \\ 

Attention Head - 2 & {\color{RANK1} \begin{CJK*}{UTF8}{gbsn}见财起意\end{CJK*} (0.480)}, {\color{RANK3} \begin{CJK*}{UTF8}{gbsn}物归原主\end{CJK*} (0.183)}, {\color{RANK4} \begin{CJK*}{UTF8}{gbsn}据为己有\end{CJK*} (0.103)}, {\color{RANK2} \begin{CJK*}{UTF8}{gbsn}偷梁换柱\end{CJK*} (0.234)}\\ \bottomrule
\end{tabular}
\caption{An example of idiom prediction with synonymic relationship. The last two rows are the attention weights assigned to the synonyms of \begin{CJK*}{UTF8}{gbsn}顺手牵羊\end{CJK*} by different heads.}
\label{tab:case}
\end{table}


As shown in Table \ref{tab:case}, the literal meaning of ground truth idiom ``\begin{CJK*}{UTF8}{gbsn}顺手牵羊\end{CJK*}" is \textit{walk off with a sheep}, which is a metaphorical way to express \textit{steal something in passing}. Our model uses 2 heads attention mechanism to assign different weights to its synonyms. In both heads, ``\begin{CJK*}{UTF8}{gbsn}见财起意\end{CJK*}" has the maximum weights, whose literal meaning \textit{seeing riches provokes evil designs} is the same with its semantics. Similarly, ``\begin{CJK*}{UTF8}{gbsn}据为己有\end{CJK*}" (literal meaning: \textit{take forcible possession of}) also has higher consistency between literal meaning and semantics. These idioms can help model comprehend ``\begin{CJK*}{UTF8}{gbsn}顺手牵羊\end{CJK*}" more efficiently. However, in head 2, model pays more attention to ``\begin{CJK*}{UTF8}{gbsn}偷梁换柱\end{CJK*}" (literal meaning: \textit{replace the beams and pillars (with inferior ones)}), which is a metaphorical idiom for describing adulterate. Our model can reduce its impact with the following gate mechanism.

\section{Related Work}
\paragraph{Machine Reading Comprehension}

Machine reading comprehension (MRC) tasks require a machine reader to answer questions with given contexts. Recently, many high-quality and large-scale reading comprehension datasets for English \cite{DBLP:conf/emnlp/RajpurkarZLL16,DBLP:conf/emnlp/Yang0ZBCSM18,DBLP:conf/naacl/DuaWDSS019} have been constructed to facilitate research on this topic. In turn, many neural models have been proposed to tackle these MRC problems, which approach or even surpass humans \cite{DBLP:conf/iclr/SeoKFH17,DBLP:conf/iclr/YuDLZ00L18,DBLP:conf/iclr/HuangZSC18,DBLP:conf/emnlp/HuPHL19}. Similarly, MRC datasets for Chinese have also been constructed, such as CMRC2018 \cite{DBLP:conf/emnlp/CuiLCXCMWH19}, DuReader \cite{HeLLLZXLWWSLWW18}, C3 \cite{SunYYC20}, etc. Among them, ChID, proposed by \newcite{chid}, is a cloze-style multiple-choice MRC dataset. It aims at Chinese idioms, also known as ``chengyu", whose non-literal and non-compositional semantic characteristics pose unique challenges for machines to understand.

\paragraph{Idiom Representation}

Good idiom representations are crucial for natural language understanding \cite{recommend,chid}. Recent works made noticeable efforts in improving the quality of idiom representations by either treating each idiom as a single token \cite{SongSLZ18,chid} or regarding idioms as regular multi-word expressions \cite{SocherPWCMNP13,WeirWRK16,QiHYLCLS19}. However, the former approaches require large-scale corpora, while the latter approaches demonstrate the essentiality of semantic compositionality \cite{pelletier1994principle}, 
which many idioms do not have.
Notably, both approaches treat idioms separately and ignore the relationship among them. Different from these studies, we utilize the synonymic relationship to obtain idiom representations. With quantitative experiments, we show that the synonymic relationship can help machines comprehend idioms better.

\paragraph{Graph Neural Network}

Graph neural network (GNN) has got extensive attention recently since they grant more flexibility to model non-Euclidean data\cite{DBLP:conf/iclr/KipfW17,DBLP:conf/nips/HamiltonYL17,VelickovicCCRLB18,DBLP:conf/icml/QuBT19}. They have shown promising performance in a range of natural language processing tasks. Our work is closely related to the graph attention networks (GATs), proposed by \newcite{VelickovicCCRLB18}. GATs can dynamically aggregate useful graph information by assigning different weights to neighboring nodes or associated edges, which are suitable for reducing the noise on the graph. In this paper, we use GATs to aggregate the synonymic relationship and instead the vanilla GATs, we incorporate the gate mechanism to further reduce the impact of incorrect synonymic relationship.

\section{Conclusion}
In this paper, we give a quantitative analysis to prove that the literal meanings of many Chinese idioms are far from their semantics, and also verify that the synonymic relationship can mitigate this inconsistency, which is beneficial for Chinese idiom reading comprehension. We propose the synonym knowledge enhanced reader to fully utilize the relationship. Experimental results show that our model achieves state-of-the-art performance among different settings of ChID, a large-scale Chinese idiom reading comprehension dataset. Similar to Chinese idioms, the inconsistency exists in a number of other language elements, e.g., slangs in English, where the potential use of the synonymic relationship among them requires a further exploration.

\section*{Acknowledgements}
We would like to thank the anonymous reviewers for their valuable feedback. This work was supported by the NSFC (No. 61936012, 61976114) and National Key R\&D Program of China (No. 2018YFB1005102).

\bibliographystyle{coling}
\bibliography{coling2020}

\end{document}